\PassOptionsToPackage{unicode=true}{hyperref} 
\PassOptionsToPackage{hyphens}{url}
\documentclass[twocolumn]{article}
\usepackage{lmodern}
\usepackage{amssymb,amsmath}
\usepackage{ifxetex,ifluatex}
\usepackage{fixltx2e} 
\ifnum 0\ifxetex 1\fi\ifluatex 1\fi=0 
  \usepackage[T1]{fontenc}
  \usepackage[utf8]{inputenc}
  \usepackage{textcomp} 
\else 
  \usepackage{unicode-math}
  \defaultfontfeatures{Ligatures=TeX,Scale=MatchLowercase}
\fi
\IfFileExists{upquote.sty}{\usepackage{upquote}}{}
\IfFileExists{microtype.sty}{%
\usepackage[]{microtype}
\UseMicrotypeSet[protrusion]{basicmath} 
}{}
\IfFileExists{parskip.sty}{%
\usepackage{parskip}
}{
\setlength{\parindent}{0pt}
\setlength{\parskip}{6pt plus 2pt minus 1pt}
}
\usepackage{hyperref}
\hypersetup{
            pdftitle={Student/Teacher Advising through Reward Augmentation},
            pdfauthor={Cameron Reid},
            pdfborder={0 0 0},
            breaklinks=true}
\usepackage{graphicx,grffile}
\makeatletter
\def\maxwidth{\ifdim\Gin@nat@width>\linewidth\linewidth\else\Gin@nat@width\fi}
\def\maxheight{\ifdim\Gin@nat@height>\textheight\textheight\else\Gin@nat@height\fi}
\makeatother
\setkeys{Gin}{width=\maxwidth,height=\maxheight,keepaspectratio}
\ifx\paragraph\undefined\else
\let\oldparagraph\paragraph
\renewcommand{\paragraph}[1]{\oldparagraph{#1}\mbox{}}
\fi
\ifx\subparagraph\undefined\else
\let\oldsubparagraph\subparagraph
\renewcommand{\subparagraph}[1]{\oldsubparagraph{#1}\mbox{}}
\fi

\makeatletter
\def\fps@figure{htbp}
\makeatother

\title{Student/Teacher Advising through Reward Augmentation}
\author{Cameron Reid}
\date{}

\begin{document}
\maketitle

\hypertarget{abstract}{%
\section{Abstract}\label{abstract}}

Transfer learning is an important new subfield of multiagent
reinforcement learning that aims to help an agent learn about a problem
by using knowledge that it has gained solving another problem, or by
using knowledge that is communicated to it by an agent who already knows
the problem. This is useful when one wishes to change the architecture
or learning algorithm of an agent (so that the new knowledge need not be
built ``from scratch''), when new agents are frequently introduced to
the environment with no knowledge, or when an agent must adapt to
similar but different problems. Great progress has been made in the
agent-to-agent case using the Teacher/Student framework proposed by
(Torrey and Taylor 2013). However, that approach requires that learning
from a teacher be treated differently from learning in every other
reinforcement learning context. In this paper, I propose a method which
allows the teacher/student framework to be applied in a way that fits
directly and naturally into the more general reinforcement learning
framework by integrating the teacher feedback into the reward signal
received by the learning agent. I show that this approach can
significantly improve the rate of learning for an agent playing a
one-player stochastic game; I give examples of potential pitfalls of the
approach; and I propose further areas of research building on this
framework.

\hypertarget{introduction}{%
\section{Introduction}\label{introduction}}

\hypertarget{reinforcement-learning}{%
\subsection{Reinforcement Learning}\label{reinforcement-learning}}

Reinforcement learning describes a variety of methods that are used to
solve sequential decision problems in which some agent is interacting
with and receiving feedback from an environment. Generally, the problem
is formulated such that the feedback is in the form of a cost (which the
agent should minimize), or a reward (to be maximized). (Sutton and Barto
2018) provide an excellent primer on the discipline of reinforcement
learning.

\hypertarget{tabular-q-learning}{%
\subsection{Tabular Q-Learning}\label{tabular-q-learning}}

The agent discussed in this paper uses a tabular Q-learning approach to
reinforcement learning described by (Watkins 1989). This is a simple but
effective approach to reinforcement learning that allows an agent to
eventually learn an exact optimal policy for every state. Put simply, we
keep a large, relatively high-dimensional table with one entry per
state-action pair; at every step, we observe a state \(S\) and take an
action \(a\) based on a policy \(\pi(s, a | Q)\). The agent then
receieves its new state \(S'\), and a reward \(R\), and makes an update
to the Q-table as follows:

\[
Q(S, a) = Q(S, a) + \alpha[R + \gamma \max_aQ(S', a) - Q(S, a)]
\]

(Sutton and Barto 2018)

where \(\gamma\) is a discount factor (lower values make the resulting
policy more ``short-sighted'') and \(\alpha\) is some step size. Under
some simple assumptions, this simple algorithm will eventually converge
to a perfectly optimal policy with probability 1 (Watkins and Dayan
1992).

\hypertarget{transfer-learning}{%
\subsection{Transfer Learning}\label{transfer-learning}}

Even the best reinforcement learning methods can be quite slow to
converge to an optimal solution. \emph{Transfer learning} describes one
attempt to solve that problem -- given an agent which has been trained
on a problem, how can we transfer the knowledge it has gained to another
agent, or generalize that knowledge to another problem?

One proposed solution in this paradigm is the Teacher-Student framework
proposed by (Torrey and Taylor 2013), in which an agent who is an
``expert'' in the problem provides advice to a learning agent to help
speed the training of the learning agent. This advice is provided by
essentially telling the learning agent which action to take at certain
states. (Da Silva, Glatt, and Costa 2017) expanded on this framework by
proposing a system in which agents learning simultaneously can all be
either advisor or advisee (or both) during training.

While those show impressive results, they make a few assumptions that I
believe are unrealistic and exhibit a departure from the fundamental
reinforcement learning problem: for example, the advice is treated as a
special case, rather than an additional environmental signal. Here, I
attempt to begin to bring those methods into harmony with the generic
reinforcement learning problem formulation.

To that end, I propose an approach for shaping the reward of a learning
agent which will help guide the learning agent as it learns by modifying
the reward received from the environment with an additional punishment
based upon the knowledge of the teaching agent.

\hypertarget{methods}{%
\section{Methods}\label{methods}}

\hypertarget{hunterprey-game}{%
\subsection{Hunter/Prey Game}\label{hunterprey-game}}

To examine this problem, I've used a simple Gridworld hunter/prey game.
The learning agent is a hunter, whose job is to catch the prey by moving
into the space occupied by the prey. The prey, in turn, moves in a
random direction at each step as long as it is not captured. The agent
is given a reward of -1 on each step that the prey has not been
captured, and a reward of 0 when it has.

The environment of the game is fully observable to the agent; at each
step, the agent receives a tuple of
\(\langle h_x, h_y, p_x, p_y \rangle\), where \(h_{\{x,y\}}\) and
\(p_{\{x,y\}}\) are the hunter's and prey's x- and y- coordinate,
respectively.

The action space in this game are integers from 0 to 3 which encode
cardinal directions in the gridworld -- 0 moves the agent down, 1 moves
the agent to the right, etc.

\hypertarget{building-the-advisor}{%
\subsection{Building the Advisor}\label{building-the-advisor}}

The advisor and agent use the same parameters for learning the policy.
First, the advisor learns a policy by playing 20,000 episodes of the
game (i.e., restarting every time the prey is captured) with the
Q-learning algorithm described above. The learning rate, \(\alpha\), was
set to 0.1; the discount factor, \(\gamma\), was set to 1.0 (so,
undiscounted learning).

The learned policy was then kept to help inform the teaching policies
described below.

\hypertarget{teaching-the-student}{%
\subsection{Teaching the Student}\label{teaching-the-student}}

Once the advisor is trained, its policy was used to provide feedback to
the learning agent using a few hand-coded policies.

In the first policy, the advisor augmented the reward the agent received
by a tunable, fixed value (set to -10 in these experiments) if the
action chosen by the agent is not the optimal action as determined by
the Q function learned by the advisor. In another policy, the teaching
agent only augments the reward signal if the chosen action is the
\emph{worst} among the four available actions. The third policy augments
the reward signal by an amount proportional to the difference between
the chosen action and the optimal action (again, as determined by the
advisor's Q function).

Note that, while these policies are fixed and provided here, the problem
of learning these policies is easily translated to a reinforcement
learning problem, and an optimal policy could easily be learned.

\hypertarget{defining-punishment}{%
\subsection{Defining Punishment}\label{defining-punishment}}

As mentioned above, I address three punishment schedules in this paper.
For simplicity's sake, I define a function \(\textbf{pun}(s,a)\) whose
value is an amount by which the reward signal is augmented before being
provided to the learning agent who has taken action \(a\) in state
\(s\). This function is defined in a few different ways, corresponding
to each of the punishment schedules.

\hypertarget{punishing-sub-optimal-actions}{%
\subsubsection{Punishing Sub-optimal
Actions}\label{punishing-sub-optimal-actions}}

The first punishment schedule imposes additional cost to agents who
choose an action which is not optimal according to the teacher's Q value
for that state. More formally, the reward is augmented by

\begin{equation}\textbf{pun}_{\text{sub}}(s,a) = C \textbf{1}_{\text{sub}}\label{eq:sub}\end{equation}

where \(\textbf{1}_{\text{sub}}\) is the indicator function whose value
is 1 when \(a \neq \max_{b}Q_{\text{teacher}}(s,b)\), and 0 otherwise.

\hypertarget{punishing-anti-optimal-actions}{%
\subsubsection{Punishing Anti-Optimal
Actions}\label{punishing-anti-optimal-actions}}

The next punishment schedule examined is similar, but only imposes an
additional cost when the learning agent chooses an action that would be
the worst among all actions. In other words

\begin{equation}
\textbf{pun}_{\text{anti}}(s,a) = C \textbf{1}_{\text{anti}}
\label{eq:anti}\end{equation}

where \(\textbf{1}_{\text{anti}} = 1\) when
\(a = \min_b Q_{\text{teacher}}(s,b)\) and 0 otherwise.

\hypertarget{continuous-punishment-by-severity}{%
\subsubsection{Continuous Punishment By
Severity}\label{continuous-punishment-by-severity}}

The final punishment schedule imposes an additional cost to the learning
agent that is proportional to the difference between the expected value
of the chosen action and the value of the optimal action based on the
teacher's action-value function. That is,

\begin{equation}
\textbf{pun}_{\text{cont}}(s,a) = C (Q_{\text{teacher}}(s,a) - \max_b Q_{\text{teacher}}(s,b))
\label{eq:cont}\end{equation}

\hypertarget{results}{%
\section{Results}\label{results}}

\begin{figure}
\hypertarget{fig:withpun}{%
\centering
\includegraphics{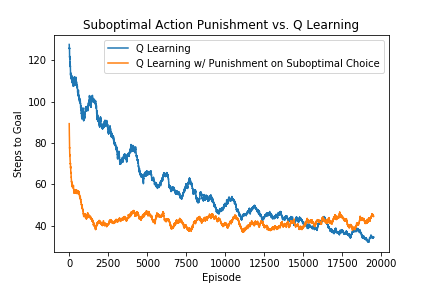}
\caption{Comparison of Q Learning and Suboptimal Action
Punishment}\label{fig:withpun}
}
\end{figure}

Figure \ref{fig:withpun} shows the results of augmenting the learning
agent's reward by Eq. \ref{eq:sub} with \(C=10\); that is, the reward
received by the agent at each step was

\begin{equation}
\hat{R}_{\text{agent}} = R_{\text{agent}} - \textbf{pun}_{\text{sub}}(s,a)
\label{eq:reward1}\end{equation}

Notice that the guidance from the teacher causes impressive improvements
in training speed; however, learning quickly levels off after only a
couple thousand episodes at a level of performance that is inferior to
what simple Q-learning achieves by the final episode. This is likely the
result of the teacher continuing to punish for slight variations to the
teacher's policy which might actually be improvements.

\begin{figure}
\hypertarget{fig:withpun2}{%
\centering
\includegraphics{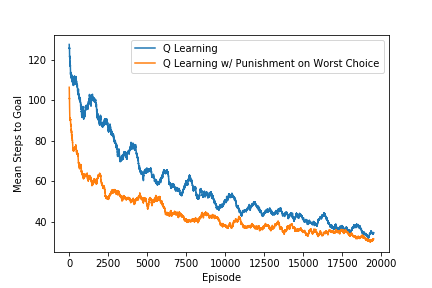}
\caption{Comparison of Q Learning and Anti-Optimal Action
Punishment}\label{fig:withpun2}
}
\end{figure}

Figure \ref{fig:withpun2} shows the results of augmenting the learning
agent's reward by Eq. \ref{eq:anti} with \(C=10\). In this case, the
reward became

\begin{equation}
\hat{R}_{\text{agent}} = R_{\text{agent}} - \textbf{pun}_{\text{anti}}(s,a)
\label{eq:reward2}\end{equation}

In this case, the increase in convergence speed was less significant;
however, because the teacher only punishes if the student chooses the
\emph{worst possible} action, the negative effects are diminished, and
the student manages to perform better than the teacher at every episode.

\begin{figure}
\hypertarget{fig:withpun3}{%
\centering
\includegraphics{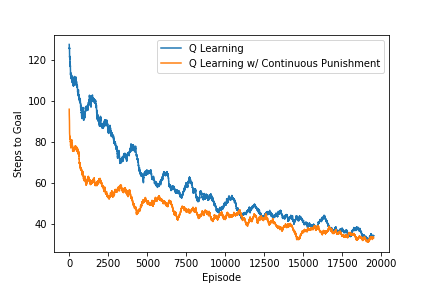}
\caption{Comparison of Q Learning and Continuous Proportional
Punishment}\label{fig:withpun3}
}
\end{figure}

Figure \ref{fig:withpun3} shows the result of augmenting the reward
signal by Eq \ref{eq:cont} with \(C=10\), i.e.,

\begin{equation}
\hat{R}_{\text{agent}} = R_{\text{agent}} - \textbf{pun}_{\text{cont}}(s,a)
\label{eq:reward3}\end{equation}

Similarly to above, it appears that, once the student has learned a
similar-enough version of the teacher's policy, the difference between
the values of the chosen action and the teacher's guess of the optimal
action are too small to make much difference, and so the student's
learning curve remains below the teacher's throughout every episode.

In this paper, I have ignored the problem of \emph{budgeting} advice,
which is prominent elsewhere in the literature. Because the continuous
punishment schedule requires advice at every step, it would certainly
require a lot of interaction with the learning agent, which most likely
makes it intractible for problems where there is a cost to interaction.
Due to this fact and the fact that the suboptimal schedule leads to poor
overall convergence, I consider the anti-optimal schedule to be the most
useful.

\hypertarget{importance-of-designing-the-feedback-policy}{%
\subsection{Importance of Designing the Feedback
Policy}\label{importance-of-designing-the-feedback-policy}}

Despite the promising results discussed above, I encountered one
instance where the reward augmentation scheme caused severe problems for
learning.

\begin{figure}
\hypertarget{fig:enc}{%
\centering
\includegraphics{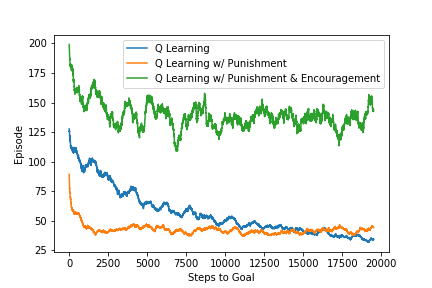}
\caption{Q-learning with Encouragement}\label{fig:enc}
}
\end{figure}

Figure \ref{fig:enc} shows the results of attempting to augment the
reward signal by providing \emph{positive} feedback when the learning
agent chooses the action that is optimal according to the teacher's Q
function, and \emph{negative} feedback as described in Eq.
\ref{eq:reward2}. Clearly, this led to a severe hindrance to training
speed, perhaps eliminating any convergence altogether; my guess in this
situation is that the agent learned to ignore the goal and instead move
to states where it had previously chosen the optimal action by chance.
Clearly, it is important to ensure that the advice provided by the
teacher is advice that will help the student learn and not create
perverse incentives.

\hypertarget{tuning-the-c-parameter}{%
\subsection{\texorpdfstring{Tuning the \(C\)
Parameter}{Tuning the C Parameter}}\label{tuning-the-c-parameter}}

Here we examine the effect of the \(C\) parameter on convergence.

\begin{figure}
\hypertarget{fig:canti}{%
\centering
\includegraphics{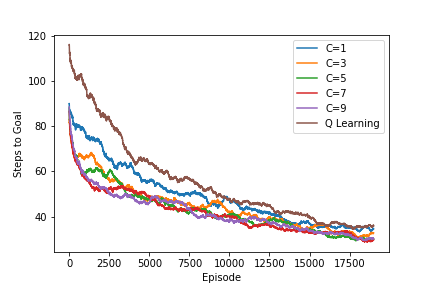}
\caption{Various \(C\)-values under the ``suboptimal''
schedule}\label{fig:canti}
}
\end{figure}

Figure \ref{fig:canti} shows the effect of tuning the \(C\) parameter in
the ``anti-optimal'' schedule. Of note is that small \(C\) values lead
to a less-impressive speedup in convergence, but result in less harmful
negative effects at later episodes. Meanwhile, a larger value of \(C\)
leads to more impressive initial speedup but leads to a policy which is
not as good. Also interesting is that performance seems to flip for
every schedule at around the 15,000 episode mark: before that, higher
values of \(C\) produce better performance but after that, higher values
of \(C\) seem to prevent further improvement.

\begin{figure}
\hypertarget{fig:csub}{%
\centering
\includegraphics{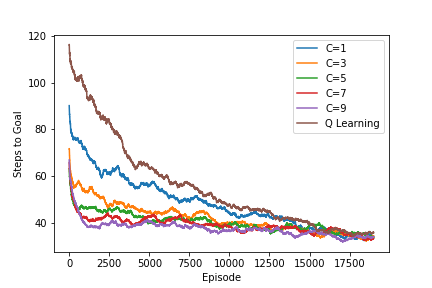}
\caption{Various \(C\)-values under the ``anti-optimal''
schedule}\label{fig:csub}
}
\end{figure}

Figure \ref{fig:csub} shows the effect of the \(C\) parameter on the
``suboptimal'' schedule. Notably, there doesn't appear to be any
negative impact on later training episodes like was apparent in Figure
\ref{fig:csub}. In fact, higher values of \(C\) seem to have purely
positive effects. Presumably, this is because the feedback is mostly
applied early in training, but once the policy becomes relatively good,
it's unlikely that the agent will take the worst possible action and so
feedback is sparse and training continues as normal.

\begin{figure}
\hypertarget{fig:ccont}{%
\centering
\includegraphics{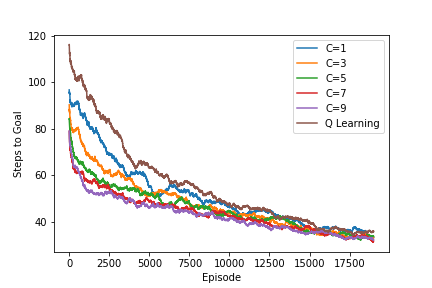}
\caption{Various \(C\)-values under the ``continuous''
schedule}\label{fig:ccont}
}
\end{figure}

Figure \ref{fig:ccont} shows the effect of tuning the \(C\) parameter
for the ``continuous'' schedule. The effect is similar to what was seen
in Figure \ref{fig:canti}: higher values of \(C\) allow the agent to
learn more quickly, but it doesn't negatively impact learning at later
epochs when the agent has learned a reasonable policy and the difference
between its chosen action and the teacher's best action is probably
small.

\hypertarget{conclusions-future-work}{%
\section{Conclusions \& Future Work}\label{conclusions-future-work}}

In this paper I've proposed an extension of the teacher/student
framework initially developed by (Torrey and Taylor 2013) which allows
the teacher to provide advice to the student via the existing structure
of reinforcement learning problems by augmenting the reward signal that
the learning agent receives from the environment. I've shown that using
this approach can significantly speed up learning. Furthermore, this
approach sidesteps some of the shortfalls of approaches like that of (Da
Silva, Glatt, and Costa 2017) -- namely, my approach extends naturally
to agents which make use of function approximation in their learning
algorithms, while the most effective approaches in (Da Silva, Glatt, and
Costa 2017) require a record of visits to a state. In problems with a
large state space where function approximation is necessary, it is
unreasonable to expect that an agent will visit any given state even two
times, so the assumption that we can count visits to a state don't hold.

The obvious next steps for this approach would be to replicate the
results of (Da Silva, Glatt, and Costa 2017) with co-learning agents.
This would require developing some metric of confidence in a state for
an agent to avoid negative impact, but that is a tractable problem.

Another obvious path forward would be to remove the hand-coded
punishment policies in favor of a trained meta-agent that learns how to
guide learning agents. This would allow not only the punishment schedule
to be optimized, but it would also lead to a formulation of the advice
budget that allows the training agent to optimize the feedback it
provides based on an actual cost of providing that feedback, again
bringing the concept of ``budget'' into the general framework of the
reinforcement learning problem.

\hypertarget{references}{%
\section*{References}\label{references}}
\addcontentsline{toc}{section}{References}

\hypertarget{refs}{}
\leavevmode\hypertarget{ref-da2017simultaneously}{}%
Da Silva, Felipe Leno, Ruben Glatt, and Anna Helena Reali Costa. 2017.
``Simultaneously Learning and Advising in Multiagent Reinforcement
Learning.'' In \emph{Proceedings of the 16th Conference on Autonomous
Agents and Multiagent Systems}, 1100--1108. International Foundation for
Autonomous Agents; Multiagent Systems.

\leavevmode\hypertarget{ref-sutton2018reinforcement}{}%
Sutton, Richard S, and Andrew G Barto. 2018. \emph{Reinforcement
Learning: An Introduction}. MIT press.

\leavevmode\hypertarget{ref-torrey2013teaching}{}%
Torrey, Lisa, and Matthew Taylor. 2013. ``Teaching on a Budget: Agents
Advising Agents in Reinforcement Learning.'' In \emph{Proceedings of the
2013 International Conference on Autonomous Agents and Multi-Agent
Systems}, 1053--60. International Foundation for Autonomous Agents;
Multiagent Systems.

\leavevmode\hypertarget{ref-watkins1992q}{}%
Watkins, Christopher JCH, and Peter Dayan. 1992. ``Q-Learning.''
\emph{Machine Learning} 8 (3-4). Springer: 279--92.

\leavevmode\hypertarget{ref-watkins1989learning}{}%
Watkins, Christopher John Cornish Hellaby. 1989. ``Learning from Delayed
Rewards.'' King's College, Cambridge.

\end{document}